\documentclass{article}

    \PassOptionsToPackage{numbers, compress}{natbib}
 \usepackage[preprint]{neurips_2026}

\usepackage{graphicx}
\usepackage[utf8]{inputenc} 
\usepackage[T1]{fontenc}    
\usepackage{hyperref}       
\usepackage{url}            
\usepackage{booktabs}       
\usepackage{amsfonts}       
\usepackage{nicefrac}       
\usepackage{microtype}      
\usepackage{xcolor}         

\usepackage{booktabs}   
\usepackage{tabularx}   
\usepackage{longtable}   
\usepackage{subcaption}
\title{Reinforcement Learning for the Game of Balatro: A Fast Simulator and Two-Policy Training Approach}

\author{%
  Ajay Vikram P\thanks{Equal contribution.} \quad
  Anjaneya Gupta\footnotemark[1] \quad
  Ethan Cheng\footnotemark[1] \quad
  Ivan Rosales\footnotemark[1] \\
  Department of Computer Science, Duke University,
  Durham, NC, USA \\
  \texttt{\{ajay.vikram, anjaneya.gupta, ethan.cheng, ivan.rosales\}@duke.edu}
}

\begin{document}

\maketitle

\begin{abstract}
Balatro, a 2024 roguelike deckbuilder, presents three difficulties that have
historically derailed reinforcement learning methods: a ${\sim}3{,}000$-action
state-dependent action space, horizons of several hundred decisions with sparse
rewards, and pervasive stochasticity in both card draws and shop generation.
Published RL results on the game remain scarce. We introduce an end-to-end
pipeline comprising (i) a high-throughput Rust simulator that is automatically
state-diffed against the live game client via BalatroBot, (ii) a condensed
observation-action interface with dynamic per-step masking of legal moves, and
(iii) a multi-agent PPO architecture that decomposes play into a tactical
\emph{blind} policy and a strategic \emph{shop} policy that share the full game
state but act in disjoint phases. Long-horizon credit
assignment is partially addressed by per-agent reward routing with explicit cross-agent credit. On the main joint run, training metrics show a sharp phase transition in which the first-blind clear rate converges above $99\%$ and mean episode return increases by roughly an order of magnitude, plateauing at a mean maximum ante of
${\sim}3$ out of $8$; single-agent baselines under identical rewards oscillated
between greedy and hoarding failure modes and did not stably clear the first
blind. These results establish the two-policy decomposition as a strong
reactive baseline for Balatro and localise the remaining gap to long-horizon
strategic planning rather than tactical blind play.
\end{abstract}

\section{Introduction}
Balatro is a single-player roguelike deckbuilder released in 2024 by LocalThunk and Playstack\cite{balatro}. A run consists of eight \emph{antes}; each ante contains three \emph{blinds} (a small blind, a big blind, and a boss blind) that the player must beat in sequence. To beat a blind, the player is dealt cards from their deck and must play poker hands until the total chip score meets or exceeds the blind's target. Between blinds, the player visits a \emph{shop} where they can buy cards, jokers that modify scoring, consumables such as tarot and planet cards, packs of cards, and vouchers that apply global effects. Money earned in the current round also generates interest, which creates a tradeoff between spending now and saving for later.

This simple outer loop hides a great deal of complexity and depth. On a tactical level, each hand of up to eight cards can be played or discarded several different ways, and the best choice depends on the exact poker hand type, the current set of jokers, whether individual cards have been enhanced or given special editions, which cards are debuffed by the boss, and how many hands and discards remain.
On a strategic level, the player is building a deck and a joker composition for the rest of the run. An early purchase that looks locally suboptimal can unlock a powerful synergy eight blinds later, and conversely, a locally tempting buy can starve the economy and cost the run at a future boss. In addition, draws, shops, pack contents, and tags are random, and many boss blinds fundamentally change the scoring rules for that one blind only.

From a reinforcement learning perspective, Balatro is attractive because it brings together three well-known sources of difficulty. The action space is very large and highly state-dependent: in a given state, many actions are illegal, and the set of legal actions can change dramatically from one step to the next. The horizon is long, with a winning run taking a few hundred decisions, and rewards are naturally sparse, where you either beat a blind or you don't. Transitions are stochastic, since the same state and action can lead to very different outcomes because of draws and shop generation. These three properties have derailed many otherwise solid RL methods in Balatro-like settings, and published results on Balatro itself remain scarce.
\section{Background}

\paragraph{Prior Balatro RL results.}
The most prominent public RL effort for Balatro is a community project by GIEWEV that reports about a 30\% simulated win rate and a 25\% real-game win rate on White Stake using a multi-agent PPO setup \cite{redditbalatro}. The author explicitly notes that their environment supports a substantial but partial subset of the full game, which is consistent with our own experience that a faithful simulator is the bottleneck. 

\paragraph{Tool-using LLM agents.}
Parallel to learning-based work, there is an emerging line of tool-calling LLM evaluation for Balatro. BalatroBot \cite{balatrobot} exposes the live game through a websocket API, letting an external agent drive real Balatro as if it were a chat environment. BalatroLLM and BalatroBench \cite{balatrollm,balatrobench} build on top of BalatroBot and evaluate LLM agents through metrics like average final round reached. These systems operate in the real game and at a much slower rate than a simulator, so they are complementary to our simulator-driven work: they cannot
easily produce the millions of rollouts that RL needs, but they are the gold standard for measuring real-game competence.

\paragraph{Game-playing RL and deckbuilders.}
Balatro sits in a broader family of card and deckbuilder domains that have attracted RL research. Community RL work \cite{slaythespire} has studied Slay the Spire, another roguelike deckbuilder with many structural similarities to Balatro (per-card effects, a shop, and long runs). Classic card RL work includes Hearthstone-style environments and single-deck optimization. Farther afield but still relevant, the body of work on MCTS plus neural networks for board games (AlphaGo, AlphaZero, MuZero) and on masked PPO in complex video games (Dota 2 \cite{openaifive}, StarCraft II \cite{alphastar}) establishes the empirical expectation that very large amounts of data and careful engineering are needed to produce human-level play, and that masking plus architecture design can dramatically reduce sample complexity.

\paragraph{Why Balatro is still unsolved.}
The public literature makes it clear that we are not the first to be humbled by Balatro. Three properties stand out as most likely to make it hard: the action space is both large and highly state-dependent, the horizon is
long, which makes credit assignment difficult and the random streams affect both in-round outcomes and between-round opportunities, so a policy can easily overfit
to a narrow slice of seed space. Every method we implement in this project was chosen, at least in part, because it is thought to help with one of these three properties.

\section{Simulator}

\paragraph{Motivation}
Current Reinforcement Learning (RL) research on Balatro generally falls into two categories, both of which present significant bottlenecks for large-scale training. The first approach, training directly on the game client, guarantees simulation fidelity but is severely throttled by the engine's frame-based execution and graphical overhead.

The second approach involves using community-developed Python "gyms." While these offer higher sampling speeds, they are often based on outdated versions of the game. Balatro has undergone significant balance patches (e.g., v1.0.1) that altered core scoring mechanics, Joker rarities, and stake scaling. Training an agent on these older environments results in "version drift," where strategies learned in the gym fail to generalize to the current live game. To reconcile the conflict between sampling throughput, logic accuracy, and version parity, we developed a custom, high-performance simulator in Rust.

\begin{figure}[t]
    \centering
    \begin{subfigure}[t]{0.48\textwidth}
        \centering
        \includegraphics[width=\linewidth]{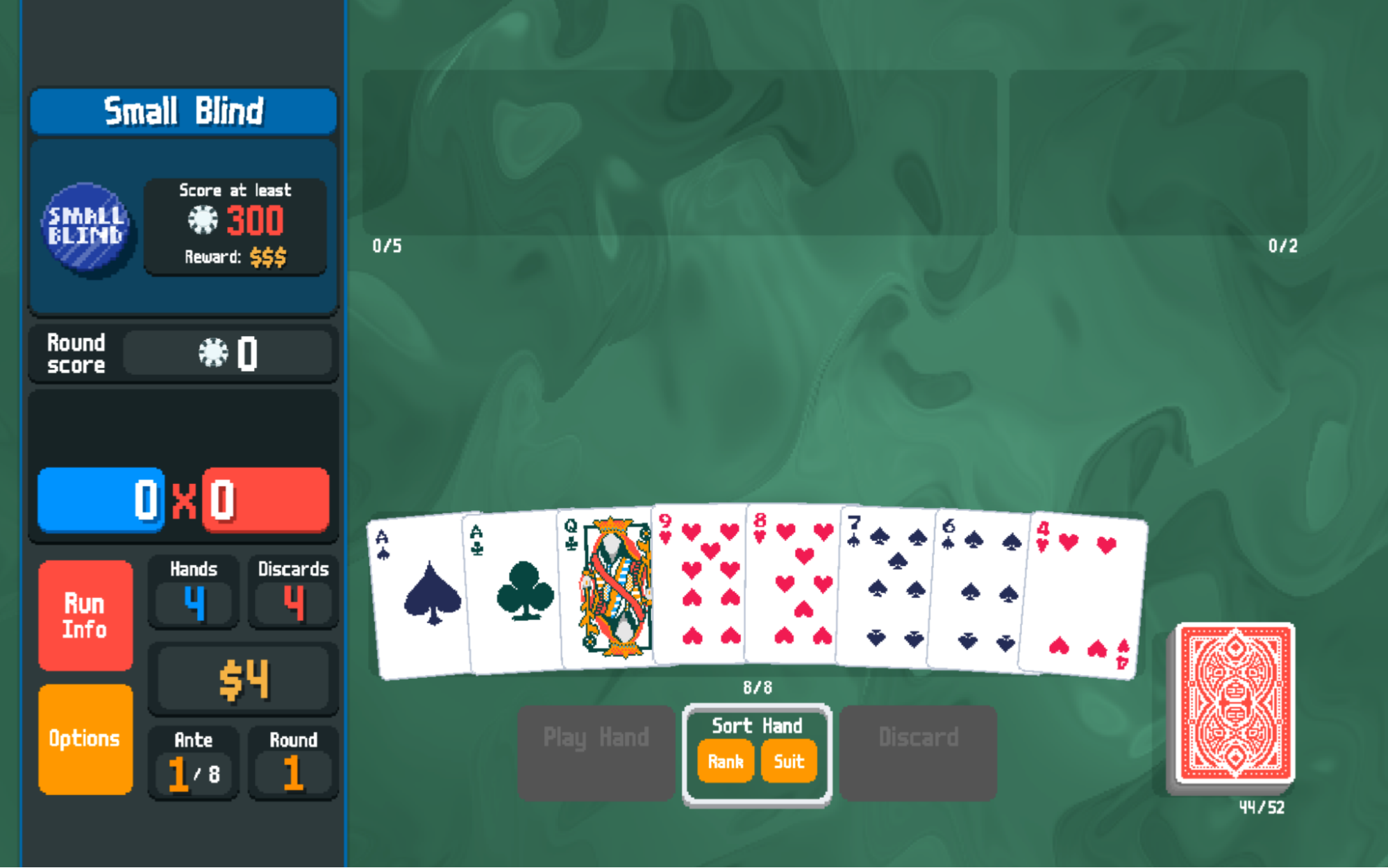}
        \caption{Original game.}
        \label{fig:original-game}
    \end{subfigure}
    \hfill
    \begin{subfigure}[t]{0.48\textwidth}
        \centering
        \includegraphics[width=\linewidth]{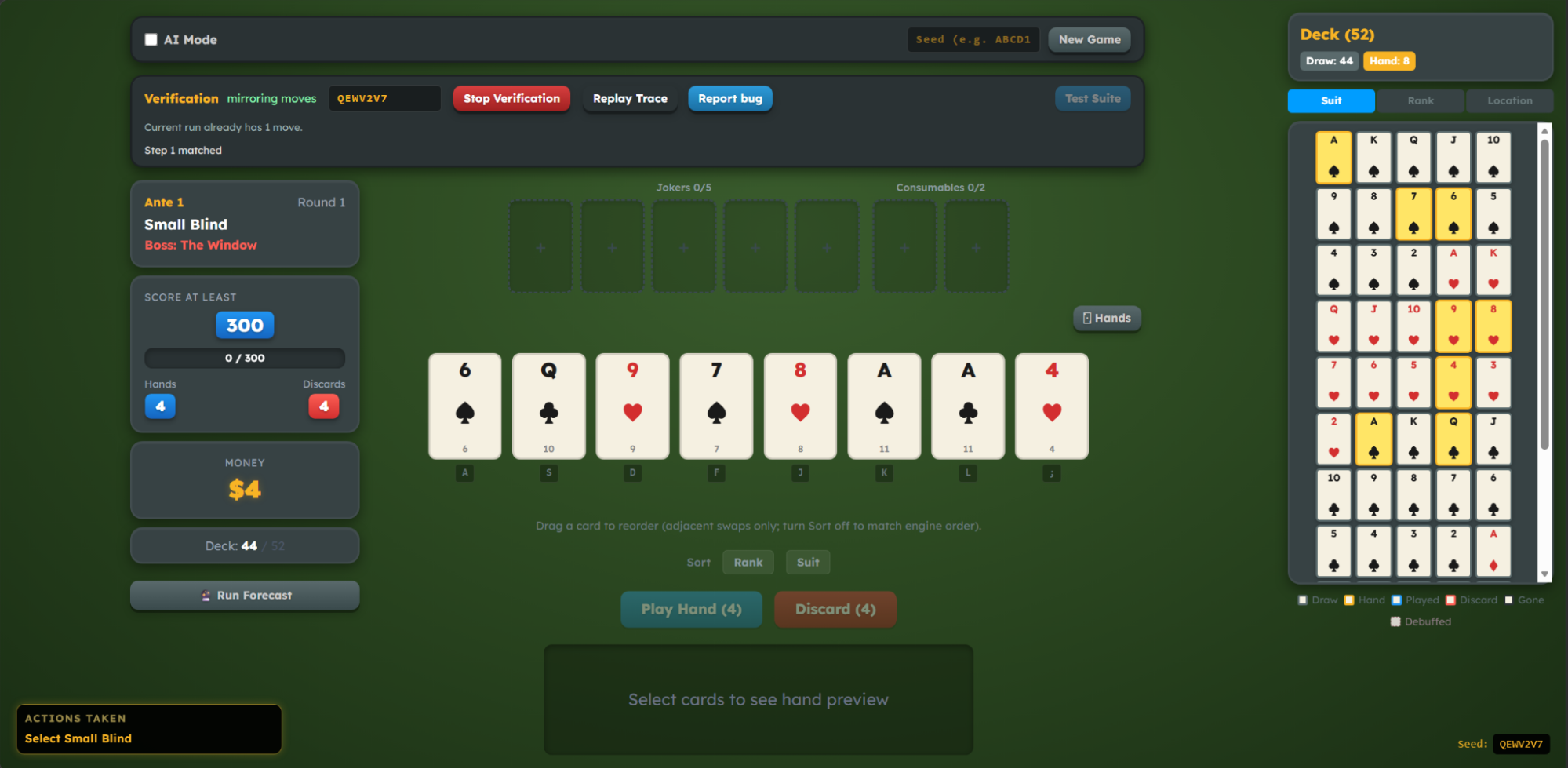}
        \caption{Our web app.}
        \label{fig:web-app}
    \end{subfigure}
    \caption{The original Balatro game (left) and our web app implementation (right).}
    \label{fig:game-comparison}
\end{figure}

\subsection{Building the Simulator}
The development of the simulator was facilitated by the extensive documentation available for Balatro’s underlying mechanics, particularly its public-facing Random Number Generation (RNG) logic. Because the game’s complexity is modular, built on discrete card effects and scoring triggers, the initial architectural setup was straightforward. We focused on replicating the scoring pipeline, ensuring that recursive triggers and held-in-hand bonuses mirrored the exact execution order and balance values found in the most recent Lua source code.

\paragraph{Python Bindings} 
The simulator's core logic is implemented in Rust for performance, but it is exposed to the training pipeline via Python bindings. This is achieved using a dedicated Rust extension, which allows high-level Python scripts to call into the high-performance Rust simulation engine.

\paragraph{Core API Methods}
\begin{itemize}
    \item \textbf{Reset:} This method initializes the environment (or a batch of environments) to a starting state. It handles the initial RNG seeding and generates the first set of observations for the agents.
    \item \textbf{Step:} This is the primary interface for progression. It takes an action and executes the underlying game logic in Rust. It then returns a vector containing the next observation.
    \item \textbf{Get State:} This provides access to the raw internal state of the simulator. It is used primarily for logging, debugging, and providing the "ground truth" during analysis that isn't directly exposed in the agent's observation space.
\end{itemize}

\subsection{Action and Observation Space} The simulator utilizes a maximal action and observation space exposed through core API methods, featuring approximately 3,000 discrete actions and a 20,000-dimensional observation space.

The architectural intuition behind this "maximalist" design was to future-proof the simulation engine. By exposing the full breadth of the game state at the API level, we created a versatile foundation that can be easily adapted for various research goals. This is achieved through a Python translation layer that wraps the core bindings, allowing us to dynamically condense or mask the state and action spaces to fit the specific scope of the model or training experiment currently being conducted. This decoupling ensures that changes to the model architecture do not require fundamental rewrites of the underlying Rust engine.

\paragraph{Action Masking}
Given the "maximalist" action space of approximately 3,000 discrete choices, many actions are invalid or contextually unavailable at any given step (e.g., attempting to play a card not currently in hand or purchasing a shop item with insufficient funds). To prevent the model from wasting training cycles on illegal moves and to simplify the policy's probability distribution, we implemented a rigorous Action Masking layer where masks are dynamically generated at each timestep based on the underlying game logic within the Rust simulator. Key constraints include:
\begin{itemize}
    \item \textbf{State-Dependent Availability:} Actions are masked according to the current game phase (e.g., masking "Play Hand" while in the Shop phase).
    \item \textbf{Resource Validation:} Actions involving gold or limited slots (like purchasing Vouchers or Jokers) are masked if the player lacks the necessary capital or inventory capacity.
    \item \textbf{Card Selection Integrity:} The mask ensures that only cards currently present in the player's hand can be selected for play or discard, effectively narrowing the 3,000-dimensional space to only those actions which are physically possible in the current state.
\end{itemize}
By providing these masks in our API methods, we ensure that the agent only samples from valid actions, significantly accelerating the convergence of the training process and ensuring the agent's behavior remains consistent with Balatro’s ruleset.

\subsection{Implementation Subsets}

While the "maximalist" engine provides comprehensive data, training directly on a 20,000-dimensional space is computationally expensive. To optimize the learning curve, we developed one specific subset of the state and action spaces. These configurations primarily manipulate the representation of hand size, deck size, joker slots, and consumable slots.
The primary challenge in representing these elements is their fluid nature; while Balatro has standard starting limits, these values can expand indefinitely through specific game mechanics. Because our observation vector uses fixed indexing to represent each card or slot, "infinite" expansion is impossible to map directly. We addressed this through two iterative configurations and padding.

\paragraph{V1: Constrained Dynamic Subset}
The objective of V1 was to significantly reduce the action space while maintaining a "buffer" for non-static game elements. This version provided a clear subset of the observation space that accounted for a small degree of variance in game state (e.g., slight increases in hand size or joker slots), allowing the model to experience some state-space expansion without the overhead of the full 20,000-dimensional vector. The specific dimensionality and mapping for this subset, including the allocation of 'buffer' indices for dynamic state expansion, are detailed in Appendix \ref{appendix:obs_mappings}.

\subsection{Verifying Simulator Accuracy}
Verification was a multi-stage process centered around a custom-built web application designed for manual visual debugging.

\paragraph{Iteration 1: Manual Synchronization via WebAssembly}
To ensure absolute parity between the training backend and the user interface, the Rust simulator was compiled to WebAssembly (WASM). This enabled the identical simulation logic to execute client-side within a browser-based debugging tool. Verification was initially performed manually: developers would mirror actions taken in the web application within the live Balatro game client to check for state divergence. While effective for initial logic testing, this process was labor-intensive, prone to human error, and hindered the pace of development.

\paragraph{Iteration 2: Automated State Verification and API Alignment}
The second iteration focused on eliminating manual overhead and ensuring strict feature parity between the simulation and the live game. The web application was updated to interface directly with the 20,000-dimensional observation space and core API methods utilized by the reinforcement learning agents, ensuring that the visual debugging tools provided an authentic representation of the model's input data.
To automate the "ground truth" comparison, we developed a verification service integrated with balatrobot. This service facilitated automated, step-by-step state synchronization between the simulator and the actual game client, allowing for the detection of subtle logic divergences across complex game-state transitions. Furthermore, we implemented a robust action-trace logging system that captured sequences of game events. These traces served two critical functions:
\begin{itemize}
    \item \textbf{Bug Replayability:} Developers could deterministically recreate and isolate edge-case failures identified during automated testing.
    \item \textbf {Regression Testing:} These traces were compiled into a comprehensive test suite, ensuring that subsequent optimizations or feature additions did not introduce regressions or undo previously implemented fixes.
\end{itemize}



\section{Core Reward Scheme}
Our Rust simulator exposes a core reward mode on top of which the per-agent shaping described in Section 5 is layered. Core is deliberately minimal and event-driven: almost all transitions return exactly zero, and the full signal comes from a small set of hand-play and terminal events. When the agent plays a hand, if the cumulative chip total meets the blind's chip target, the blind is cleared and that step returns a reward of 1. If the blind is not yet cleared but hands remain, the step returns the chip total from that single play divided by the blind's chip target, capped at 1. This gives a dense per-hand signal proportional to the fraction of the blind that one play covered, with the cap keeping reward magnitudes bounded under PPO's clipping. Running out of hands without clearing the blind terminates the episode with a reward of -1, as does a stalemate in which no legal action exists in a non-terminal state. Actions that are out of range or rejected by the legal-action mask also return -1 without forcing termination; in practice the masking layer prevents these transitions from ever reaching the environment, so this branch functions as a safety net rather than a live training signal. All remaining transitions, including entering and leaving the shop, selecting or skipping blinds, discarding, cashing out, rerolling, and reordering jokers or hand cards, return exactly zero in the core scheme. The scalar therefore credits neither money nor deck composition directly; both matter only insofar as they enable a later hand to score. This design keeps the base reward a clean, low-variance backbone: every denser signal in our experiments, including hand-type quality, discard credit, urgency, graduated loss penalty, interest-tier preservation, joker growth, and cross-agent routing, is added in the Python wrapper on top of the core scheme, which lets us ablate the shaping layer against the event-based base without touching the simulator.

\section{Multi-agent RL}
\begin{figure}[t]
  \centering
  \includegraphics[width=\linewidth]{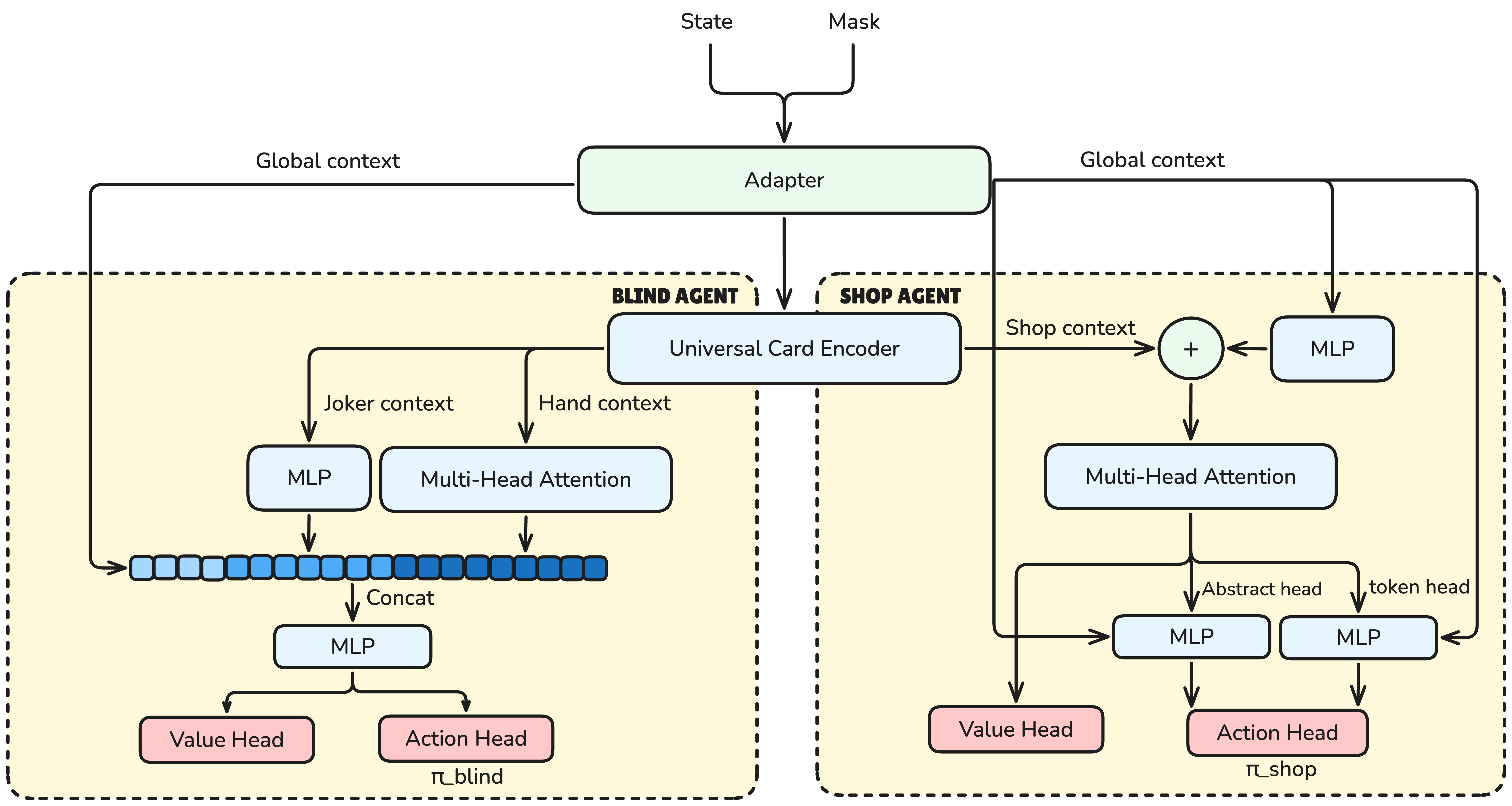}
  \caption{Multi-agent setup for blind play and shop decisions: shared observation preprocessing and a universal card encoder feed agent-specific backbones; the blind policy attends over hand (and joker) tokens before a joint trunk and heads, while the shop policy injects a projected global state into action tokens, applies attention, and merges token-level and abstract (meta-action) logits into the full discrete action space.}
  \label{fig:multi-agent-architecture}
\end{figure}

The multi-agent pipeline is the focus on our RL work. This section describes why we split the game into two policies - the blind policy and the shop policy, how each policy is designed, how they are trained together, and what the learning curves were like. Figure~\ref{fig:multi-agent-architecture} summarizes the architecture.

\subsection{Motivation: why two policies?}
In early single-agent runs, we observed that the policy oscillated between greedy play behavior, in which it would commit cards aggresively and ignore long-term economy, and hoarding behavior in which it would rarely buy anything in the shop and accumulate money without compounding synergies. Neither behavior is optimal, and the oscillation suggested that the same parameters were trying to amortize two very different skills.

Play inside a blind is a short, tactical problem where the agent sees a small number of cards, decides on a poker hand, and commits. The decision is mostly local with respect to which cards are already selected, what does each card contribute, how many hands and discards remain. Shop play is structurally different in the sense that the agent sees prices, items, and jokers, decides how to spend money, and must reason about compounding synergies with what it already owns. It has a much longer causal reach, and depends on economy-management skills that the blind policy does not teach. 

This led us to split the agent into two cooperating specialist policies. Both policies share the full game state (nothing is hidden from one and shown to the other) but each only acts during the phases it is responsible for while receiving shared reward signal for the outcomes that matter, such as winning a blind or a run. Each agent has its own feature extractor that reshapes and encodes the raw observation differently before feeding it into its neural network.

\subsection{The blind policy}

\paragraph{Inputs} The blind policy is invoked whenever the game is in the "Selecting a Hand" phase. It receives the compact observation and an adapter breaks it into logical pieces - a sequence of up to ten hand cards with their per-card properties, a sequence of up to eight jokers with their editions and sell values, a short vector of blind and round scalars such as chips scored, chips needed, hands remaining, and discards remaining, and a small auxiliary vector describing which poker hand types are currently available given the selected cards.

\paragraph{Encoding} We employ a \textit{UniversalCardEncoder} module that turns raw card data into learned vector representations. Each card is turned into a token by summing several learned embeddings: one for the card's identity, one for its suit, one for its rank, plus embeddings for its enhancement, edition, and seal, along with a debuff flag. The suit and rank embeddings also include relational features computed on-the-fly: how many other cards share the same rank, same suit, whether a flush is possible, whether adjacent ranks exist (straight detection), and sinusoidal rank encodings. Both encoders share the same weights, so they develop a common understanding of what each card means.

\paragraph{Representation} After encoding, the hand cards plus a special context token go through a transformer encoder with multi-headed self-attention. Attention is a natural choice here given that the best action is often dependent on pairs or small groups of cards and self-attention is designed to exploit that kind of relational structure. The output of the special context token is used as a single dense summary of the hand. In parallel, the joker tokens are pooled through a small MLP into a joker summary. This deliberate asymmetry reflects how the game works - decisions in a blind are dominated by which cards to commit right now, while jokers provide a relatively stable multiplier structure that is better summarised than attended over at every step. The hand summary, joker summary, and a summary vector for the non-card features are concatenated and fed through a small MLP backbone. The output of this backbone is the representation used by both the policy head and the value head.

\paragraph{Action head} The policy head is not a single flat softmax over the entire action space. Instead, it is autoregressive: the agent first decides whether it intends to play or discard, and then, conditioned on that intent and on the cards currently selected, it repeatedly decides either to add another card to the selection or to stop and commit. The motivation for this structure is that the action "commit a five-card flush" is not the same kind of decision as "toggle the seventh card." Autoregressive heads let the model express a joint preference over a set of card selections plus a commit action without exploding the action space. In practice, they also train more stably on the blind policy than a flat head did in our early experiments.

\paragraph{Value head} The value head is a small two-layer MLP on top of the backbone representation. It predicts the expected discounted return from the current state, and is used as the critic in PPO.

\subsection{The shop policy}

\paragraph{Inputs} The shop policy is invoked whenever the game leaves a blind -- in the shop, when opening a pack, and at the blind-select prompt. The compact observation is sliced into four logical groups: a \emph{state summary} (money, round info, deck statistics, a learned blind embedding, joker count and capacity, and per-hand-type scoring stats), the \emph{items currently for sale} (shop cards, packs, and vouchers together with their prices), the items the agent currently \emph{owns} (jokers, consumables, and hand cards with their sell values), and the items inside an \emph{opened pack} when applicable. 

\paragraph{Encoding} Every item, whether it is in the shop, in the pack, or already owned, is turned into a token by the same \textit{UniversalCardEncoder} used by the blind policy, so jokers, playing cards, and consumables all share a common representation across the two agents. Each token is additionally stamped with a small action tag (buy, sell, use, move, pack-select) and a slot id, so the network can tell, for example, the "buy" version of a joker apart from the "sell" version of that same joker. A learned \emph{shop-context} token is prepended to the sequence to play the role of a CLS token, and a projection of the flat state summary is broadcast-added to every token so each item is evaluated with the current money, blind, and board in mind.

\paragraph{Representation} The full sequence, shop-context token followed by all item tokens, is processed by a stack of multi-head attention layers. Attention is again a natural fit: the value of a shop item is rarely intrinsic and almost always depends on what the agent already owns (a multiplier joker is strong only if the board can produce chips, a scoring-card joker pairs well with already-owned enhancements, and so on). The updated shop-context token then serves as a single dense summary of and is used by both the action and value heads. The updated item tokens are used to score item-specific actions.

\paragraph{Action head} The shop's action space has two qualitatively different families, and the head treats them separately. Item-level actions (buying a specific card, selling a specific joker, choosing a specific pack item, \ldots) are scored through an \emph{intent} mechanism: a small MLP reads the shop-context token and state summary and produces a single "intent vector" that captures what the agent currently wants, and each item's logit is the dot product of this intent with that item's attended embedding. Because the same item always produces the same embedding, this scoring is naturally position-invariant - the policy learns "I want this joker" rather than "I want slot 0," which is where a flat softmax over action indices struggles. Abstract actions that are not tied to any item (rerolling the shop, ending the shop, selecting or skipping the next blind, cashing out, rerolling the boss) are scored by a separate small MLP on top of the shop-context token, because giving them their own parameters worked better than treating them as pseudo-item tokens. The two heads are then stitched back into a single flat logit vector and illegal actions are masked with $-\infty$ before the softmax, exactly as in the blind policy.

\paragraph{Value head} The value head is a small MLP on top of the \emph{same} shop-context token that drives the action head. Sharing this representation between actor and critic ties the policy's "what do I want in this shop" signal to the critic's "how good is this shop state" signal through a common latent, which we found to stabilise shop training noticeably.

\subsection{Per-agent rewards}

Plain PPO with a single environment reward tends to assign credit poorly in a multi-agent setup. We therefore layer per-agent shaping on top of the base reward. The shaping is implemented as a routing step: given the raw reward from one environment step, we decide how much of it to route to each agent, and we add agent-specific bonuses that reward agent-appropriate behaviour.

For the blind agent we route the core step reward to it when it is acting, and we add six possible dense signals. First, a hand-type quality bonus that rewards playing higher-tier poker hands more than lower ones, even when chip gains are small. Second, a discard-credit bonus that gives the agent partial credit when a hand that contained a discard ends up making chip progress, so the agent is not discouraged from discarding when discarding is the right call. Third, an urgency term that scales reward by how few hands remain, pushing the agent to score more when the game is close to failing. Fourth, a graduated loss penalty that reduces the punishment for losing a run proportionally to how far the agent got, so an agent that makes it to ante four is not punished as harshly as one that dies in ante one. Fifth, an ante-scaled clear bonus so that clearing a later blind is worth more than clearing an early one. Sixth, an efficiency bonus when a blind is cleared with hands to spare.

For the shop agent we add two dense signals. An interest-tier preservation penalty subtracts from reward whenever the shop agent spends enough money to cross a five-dollar interest boundary, so it internalises the opportunity cost of lost interest. A joker-growth bonus rewards increasing the number of owned jokers, because in practice building a joker-heavy board is almost always a good strategic move that the early agent is otherwise slow to learn.

Cross-agent signals are equally important. When the blind agent clears a blind, the cash-gained reward and round-won reward are credited to the \emph{shop} agent, because the shop agent is what will spend the cash next. When the run is won, both agents receive the full run-win bonus, because it is a shared outcome. This kind of explicit cross-agent credit assignment is important to keep the shop agent active in the learning process.

\subsection{Training}
Both agents are trained jointly using Ray RLlib's multi-agent PPO implementation. The environment wrapper runs a single game instance per worker and routes each observation and reward to either the blind or shop agent based on the current phase; the two policies are stored and updated as distinct RLlib policies
with independent parameters but a shared value-loss budget. A typical run uses 4 env-runner processes ($1$ environment each) feeding a single NVIDIA
RTX A5000 learner, which we found to be the sweet spot for an A5000: more runners saturate the shared GPU, fewer starve the learner.

\paragraph{PPO hyperparameters.} Unless stated otherwise, every run uses the defaults in
Table~\ref{tab:ppo-hparams}. Each training iteration collects
$5{,}120$ environment steps, which are then reused for
$4$ SGD epochs at a minibatch size of $512$ (so $\sim\!40$ gradient updates per
iteration). The learning rate is a flat $3\times 10^{-5}$, the clipping threshold is
$0.15$, the value-function clip is $3.0$ with a loss coefficient of $0.25$, and we apply
a global gradient-norm clip of $3.0$. We use $\gamma=0.99$ for both agents. Advantages
are computed with RLlib's default GAE.

\paragraph{Entropy schedule.} The two agents use different exploration pressures
because their effective branching factors are very different. The blind agent uses a constant entropy coefficient of $0.001$ throughout. The shop agent starts at $0.02$ and decays linearly over environment steps: $0.02 \to 0.006$ by $1.5\mathrm{M}$ steps, and $0.006 \to 0.001$ by $4\mathrm{M}$ steps. Without this schedule the shop policy either collapsed to "reroll forever" or "end shop immediately" within the first few hundred iterations.

\paragraph{Scale of our longest run.} Our main blind--shop run trains for
$4{,}000$ PPO iterations, i.e.\ approximately $20.5\mathrm{M}$ environment steps total, with checkpoints saved every $10$ iterations. Deterministic evaluation uses $32$ episodes per evaluation pass with exploration disabled. Shorter sweeps and ablations were run at $300$--$500$ iterations under identical PPO settings.

\begin{table}[t]
\centering
\small
\caption{PPO training hyperparameters used for the joint blind--shop run.}
\label{tab:ppo-hparams}
\begin{tabular}{ll}
\toprule
\textbf{Hyperparameter} & \textbf{Value} \\
\midrule
Env-runner processes ($\times$ envs per runner) & $4 \times 1$ \\
Train batch (env steps / iteration)             & $5{,}120$ \\
SGD minibatch size                              & $512$ \\
SGD epochs per iteration                        & $4$ \\
Rollout fragment length                         & auto (\texttt{truncate\_episodes}) \\
Discount $\gamma$                               & $0.99$ \\
PPO clip parameter                              & $0.15$ \\
Value-function loss coefficient                 & $0.25$ \\
Value-function clip parameter                   & $3.0$ \\
Gradient-norm clip                              & $3.0$ \\
Learning rate                                   & $3\times 10^{-5}$ (flat) \\
Blind entropy coefficient                       & $0.001$ (flat) \\
Shop entropy coefficient                        & $0.02 \to 0.006 \to 0.001$ (by env steps) \\
Training iterations (main run)                  & $4{,}000$ \\
Total environment steps (main run)              & ${\sim}20.5\mathrm{M}$ \\
Checkpoint frequency                            & every $10$ iterations \\
Evaluation episodes / pass                      & $32$, deterministic (\texttt{explore=false}) \\
Hardware                                        & single RTX A5000 GPU \\
\bottomrule
\end{tabular}
\end{table}

\subsection{Analysis}

\begin{figure*}[t]
    \centering

    \begin{subfigure}[t]{0.31\textwidth}
        \centering
        \includegraphics[width=\linewidth]{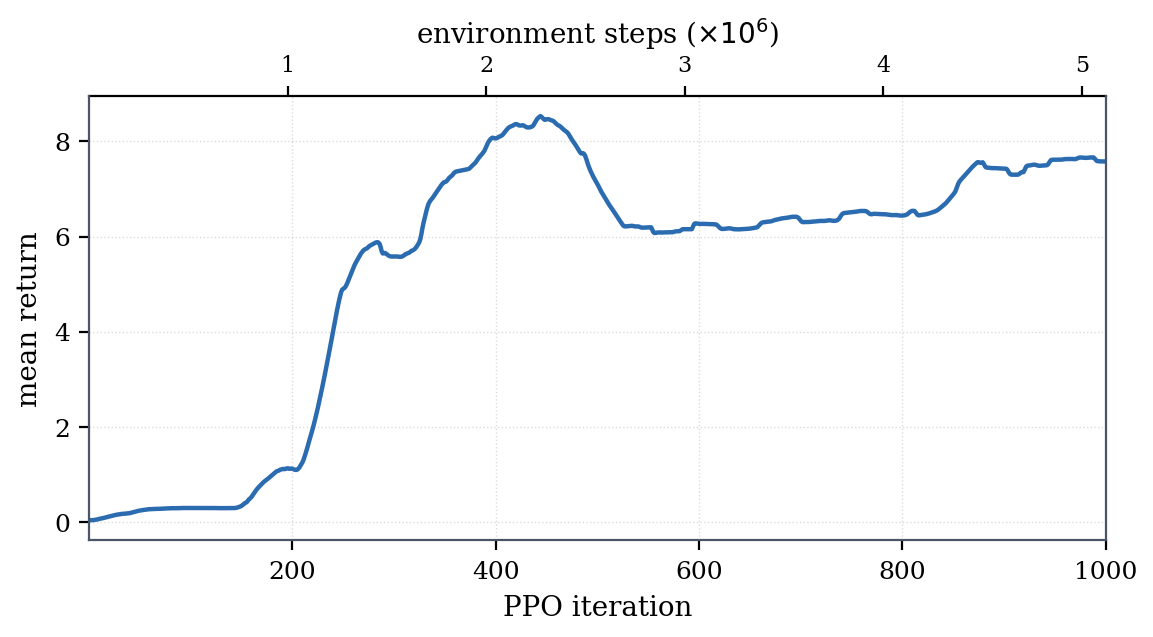}
        \caption{Mean episode return.}
        \label{fig:metric-return}
    \end{subfigure}
    \hfill
    \begin{subfigure}[t]{0.31\textwidth}
        \centering
        \includegraphics[width=\linewidth]{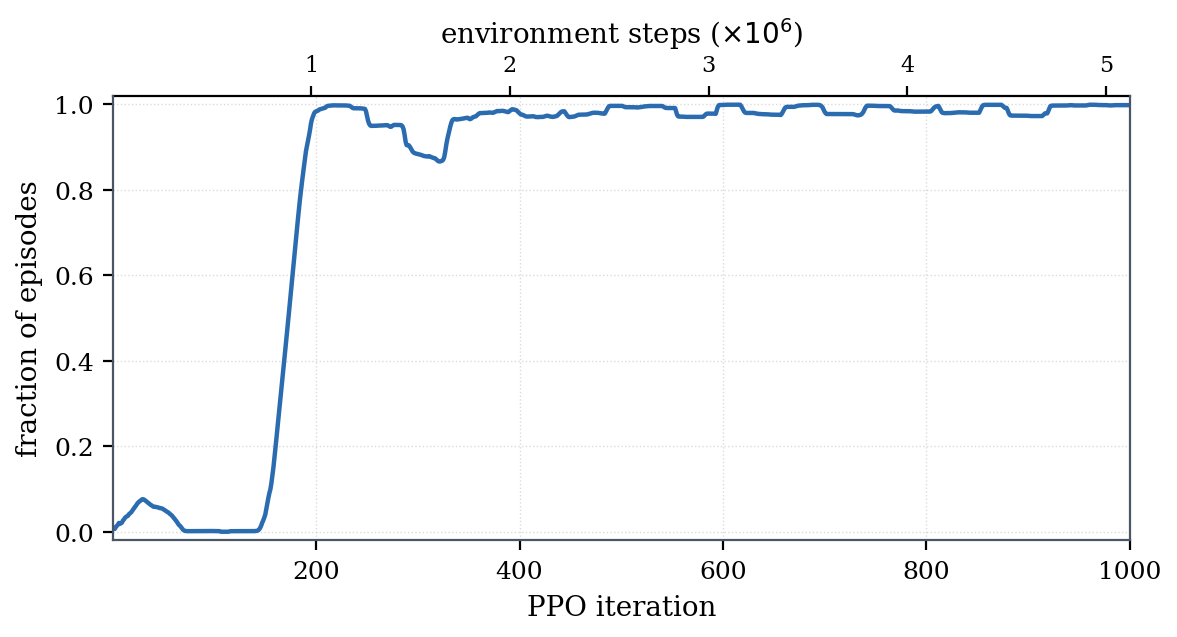}
        \caption{First-blind clear rate.}
        \label{fig:metric-blind-clear}
    \end{subfigure} 
    \hfill
    \begin{subfigure}[t]{0.31\textwidth}
        \centering
        \includegraphics[width=\linewidth]{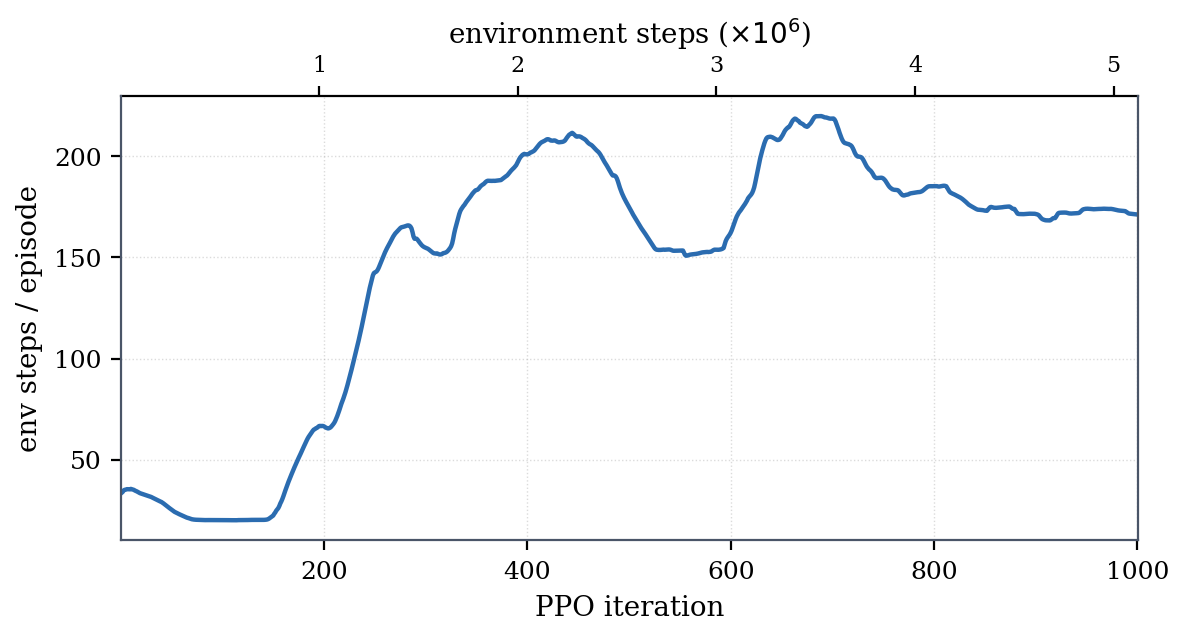}
        \caption{Mean episode length.}
        \label{fig:metric-ep-len}
    \end{subfigure}

    \vspace{0.8em}

    \begin{subfigure}[t]{0.31\textwidth}
        \centering
        \includegraphics[width=\linewidth]{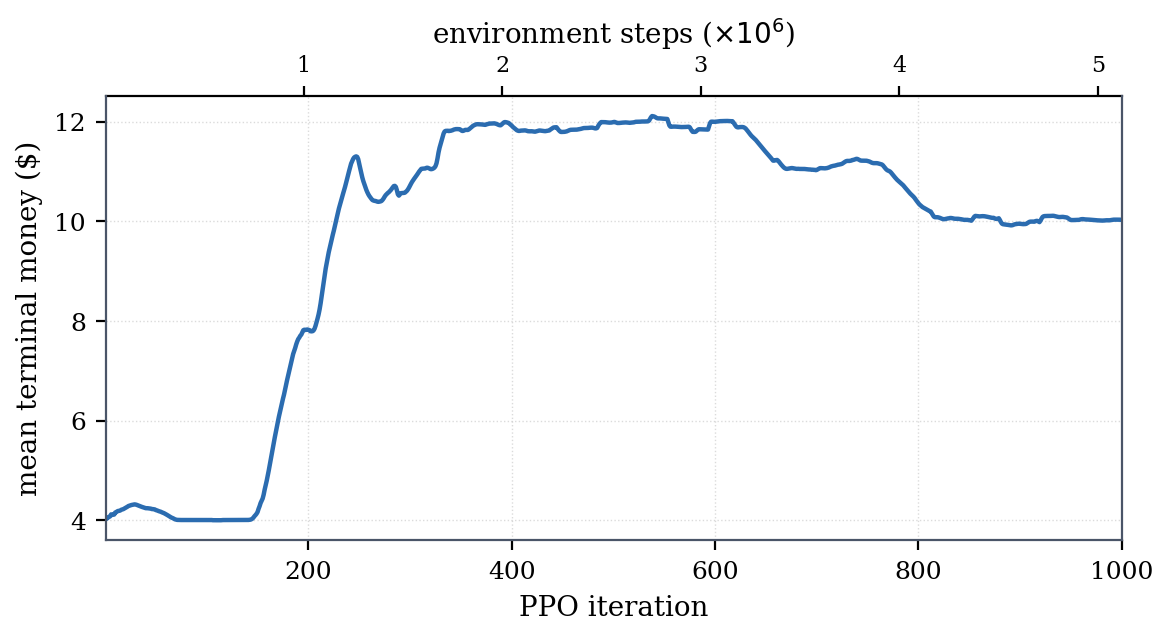}
        \caption{Mean terminal money.}
        \label{fig:metric-money}
    \end{subfigure}
    \hspace{0.035\textwidth}
    \begin{subfigure}[t]{0.31\textwidth}
        \centering
        \includegraphics[width=\linewidth]{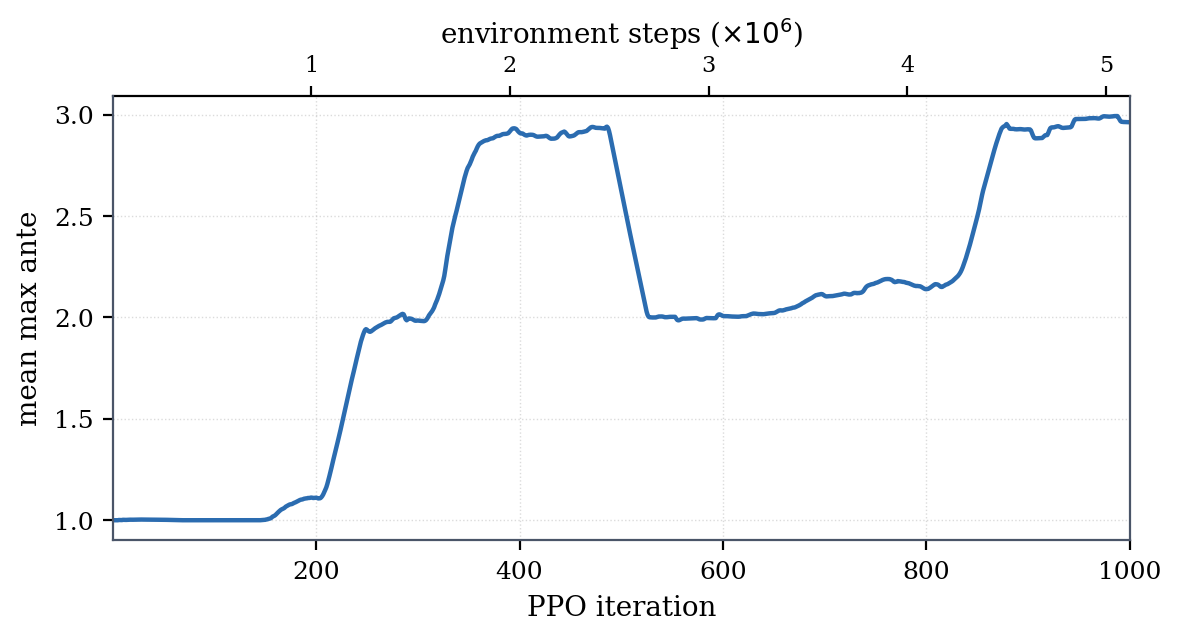}
        \caption{Mean maximum ante reached.}
        \label{fig:metric-ante}
    \end{subfigure}

    \caption{Training metrics for the joint MARL run over the first 1000 PPO iterations. Curves are smoothed using a rolling mean with window size 40.
    }
    \label{fig:training-metrics-grid}
\end{figure*}

Figure~\ref{fig:training-metrics-grid} shows five training-time metrics over the first
$1{,}000$ PPO iterations of the joint blind--shop run, each smoothed with a trailing
$40$-iteration rolling mean. All curves are computed over the on-policy training batch
with exploration enabled, so they reflect behaviour under the current stochastic policy
rather than deterministic evaluation; we discuss the gap in the limitations below.

\paragraph{A three-regime learning trajectory.}
The five panels tell a consistent story organised into three regimes. For roughly the
first $150$ iterations, the joint policy is in a low-skill regime: mean episode return
stays under $1$, the first-blind clear rate hovers near zero with long stretches of
exactly $0.0$, episodes are short ($\sim\!20$--$27$ steps), terminal money sits at the
starting bankroll of $\$4$, and max ante reached is essentially locked at $1$. Episodes
are terminating at or before the first blind, and no shop-level behaviour can yet be
learned because the shop policy rarely gets to act with meaningful state.

Between iterations $\sim\!150$ and $\sim\!400$, all five metrics move together in what is
best described as a single phase transition rather than gradual improvement. The
first-blind clear rate crosses $0.5$ at iteration $151$, $0.9$ by $160$, and $0.99$ by
$173$; over the same window mean return jumps by roughly an order of magnitude (segment
mean $\sim\!5.2$, peaks near $9$), mean episode length rises from $\sim\!25$ to
$\sim\!150$ steps, terminal money moves from $\$4$ to $\sim\!\$11$, and mean max ante
climbs from $1$ to $\sim\!2.1$. The simultaneity is not coincidental: once the blind
policy reliably passes the first gate, the shop policy begins to see varied, well-funded
states, and the extra episode length is causally downstream of clearing blinds rather
than an independent effect.

From iteration $\sim\!400$ onward the system enters a plateau. Smoothed return settles
around $7.6$, blind clear rate at $\sim\!0.998$, episode length in the $170$--$200$ step
band, terminal money near $\$10$--$\$12$, and mean max ante near $\sim\!2.6$--$3.0$.
Within-band fluctuations are consistent with shop RNG, pack and voucher variance, and
differing joker-build trajectories rather than with further improvement in blind
competence.

\paragraph{What the curves do and do not establish.}
Jointly, the metrics support a specific claim: the two-policy decomposition with routed
per-agent rewards successfully learns the first-blind sub-problem and makes the
multi-phase game tractable for downstream learning. They do \emph{not} establish that
the agent wins runs. A full Balatro run spans eight antes; our mean max ante reached
during training is $\sim\!3$, with batch-level maxima of $3$ in the later regime. In
other words, the joint system is competent at the tactical blind-play subtask and is
making structural progress through the roguelike loop, but it is not yet a run-completing
policy. We view this as the expected outcome of the scale and curriculum used here
($\sim\!5\mathrm{M}$ steps at this checkpoint, no explicit ante curriculum, no search).

\paragraph{Training vs.\ deterministic evaluation.}
Because these are on-policy training rollouts, each per-iteration value averages over a
finite batch of stochastic episodes and includes entropy-driven exploration; this both
inflates variance and can bias metrics relative to a clean $\texttt{explore=false}$ eval.
Under deterministic evaluation the shop policy's entropy schedule no longer injects
noise into purchases, which tends to shift blind clear rates upward but max-ante means
only modestly, consistent with the interpretation that late-game progression is bottlenecked
by strategy rather than by exploration noise.

\paragraph{Caveats.}
Three caveats are worth stating explicitly. First, we plot only the first $1{,}000$ of
$4{,}000$ iterations here because the phase-transition structure is the main qualitative
finding; return and clear rate continue to drift within their plateau bands over the
remaining $3{,}000$ iterations but do not change regime, and mean max ante rises only
marginally. Second, the rolling-mean window ($40$) smooths the blind-clear transition,
so the smoothed curve lags the raw crossing by roughly $20$--$40$ iterations. Third, the
five metrics are correlated rather than independent: episode length rises because clears
succeed, terminal money rises because episodes reach the shop, and so on. We therefore
read them as a joint diagnostic of the two-policy architecture rather than as five
separate tests.

\section{Limitations}

Several limitations bound the claims in this paper. (i) \emph{Simulator coverage.} Our
V1 observation/action subset does not expose the full joker and voucher pool and caps
dynamic slots at fixed buffers; strategies that depend on late-game legendary jokers or
on stacking beyond the buffer are out of reach by construction. (ii) \emph{Scale.} At
$\sim\!20.5\mathrm{M}$ environment steps the main run is one to two orders of magnitude
short of compute budgets typical for agents that master complex, stochastic games; we
view our results as evidence of learnability rather than of mastery. (iii) \emph{Seed
generalization.} We do not evaluate against held-out seed distributions or against the
live game client, so train--eval gaps and overfitting to simulator-specific RNG cannot
be ruled out. (iv) \emph{No search.} Our policies are purely reactive; any claim about
the hardness of specific blinds or builds is conditional on not using lookahead at
decision time. (v) \emph{Single training run for the main result.} We report one main
run with ablations at shorter horizons; seed variance across full-length runs is not
characterised.

\section{Conclusion}

We reported a first end-to-end reinforcement learning pipeline for Balatro that combines a high-throughput Rust simulator with automated parity checks against the live client, a maximal action and observation interface with dynamic action masking, and a two-policy multi-agent PPO system in which a tactical blind policy and a strategic shop policy are trained jointly over a shared game state with per-agent reward routing. Training metrics on the first $5\mathrm{M}$ environment steps of the main joint run show a clean three-regime learning trajectory: an initial low-skill regime in which episodes fail at the first blind, a sharp phase transition during which the blind policy converges to $>\!99\%$ first-blind clear rate and episodes lengthen by roughly an order of magnitude, and a plateau in which the system plays economically meaningful but not yet winning runs, reaching a mean maximum ante of $\sim\!3$ out of $8$. The decomposition of the agent into two cooperating specialists with shared observations but agent-appropriate rewards is, in our experiments, what made this trajectory achievable; single-agent baselines oscillated between greedy and hoarding modes and did not stably clear early blinds. The main open gap is the long-horizon strategic layer: our policies solve the tactical sub-problem but do not yet complete full runs.

\section{Future Directions}
While our current experiments focused on the feasibility of PPO-based models, our findings suggest that a more specialized approach is required to master the complex, sequential decision-making of Balatro. A significant pivot for future work involves moving beyond model-free algorithms like PPO in favor of Monte Carlo Tree Search (MCTS)-based methods, specifically AlphaZero and MuZero. The integration of MCTS would provide the "look-ahead" capabilities necessary to evaluate the long-term scoring potential of specific card combinations, a depth that PPO struggled to achieve. Finally, continuous refinement of the Rust simulation engine and the automated verifier remains a top priority. Future development will focus on expanding state coverage to include complex interactions such as legendary jokers and exotic vouchers that were previously out of scope. We also aim to enhance the balatrobot verification service to include asynchronous state checks, ensuring the simulator maintains a 100\% faithful representation of the live game client across all possible seeds.

Another promising direction is to treat Balatro as a reasoning task for a language model. We would first have a strong teacher model like Gemini 3 Pro play through a handful of runs to produce a small corpus of high-quality gameplay traces, then fine-tune a smaller open model on these traces as a warm-up, and finally refine it with Group Relative Policy Optimization (GRPO) using the game score as reward. This recipe has already shown strong results on similar multi-step card games, and it fits Balatro well: joker synergies and shop purchases are the kind of decisions that are naturally argued about in words, and starting from a competent teacher policy avoids the sparse-reward exploration problem we ran into with PPO.

\appendix
\section{V1 Action and Observation Space Mappings}
\label{appendix:obs_mappings}

\begin{table}[ht]
\centering
\small
\caption{V1 action space mapping (74 discrete actions).}
\label{tab:action-space}
\begin{tabularx}{\textwidth}{@{}llX@{}}
\toprule
\textbf{Range} & \textbf{Count} & \textbf{Action} \\
\midrule
0--9   & 10 & \texttt{SELECT\_CARD\_}$i$ \\
10     & 1  & \texttt{PLAY\_HAND} \\
11     & 1  & \texttt{DISCARD\_HAND} \\
12--15 & 4  & \texttt{USE\_CONSUMABLE\_}$i$ \\
\addlinespace
16--25 & 10 & \texttt{BUY\_}$i$ — shop slot $i$, indexed to match \texttt{shop\_prices} \\
26     & 1  & \texttt{REROLL\_SHOP} \\
27     & 1  & \texttt{END\_SHOP} \\
\addlinespace
28--35 & 8  & \texttt{SELL\_JOKER\_}$i$ \\
36--39 & 4  & \texttt{SELL\_CONSUMABLE\_}$i$ \\
40     & 1  & \texttt{SELECT\_SMALL} \\
41     & 1  & \texttt{SELECT\_BIG} \\
42     & 1  & \texttt{SELECT\_BOSS} \\
43     & 1  & \texttt{SKIP\_BLIND} \\
\addlinespace
44--48 & 5  & \texttt{PACK\_SELECT\_}$i$ \\
49     & 1  & \texttt{PACK\_SKIP} \\
50     & 1  & \texttt{CASH\_OUT} \\
51     & 1  & \texttt{REROLL\_BOSS} \\
\addlinespace
52--59 & 8  & \texttt{MOVE\_JOKER\_RIGHT\_}$i$ \\
60--63 & 4  & \texttt{BUY\_AND\_USE\_}$i$ — shop slots 0--3 only (zero-target consumables) \\
64--73 & 10 & \texttt{MOVE\_HAND\_RIGHT\_}$i$ — swap hand slot $i$ with $i{+}1$ \\
\bottomrule
\end{tabularx}
\end{table}

\begin{table}[ht]
\centering
\small
\caption{V1 observation space mapping (1449 dimensions). Left and right halves are
         contiguous; split here for typesetting.}
\label{tab:obs-space}
\begin{minipage}[t]{0.48\textwidth}
\centering
\begin{tabular}{@{}lll@{}}
\toprule
\textbf{Range} & \textbf{Count} & \textbf{Field} \\
\midrule
0--9        & 10  & \texttt{hand\_cards} \\
10--19      & 10  & \texttt{hand\_enhancements} \\
20--29      & 10  & \texttt{hand\_editions} \\
30--39      & 10  & \texttt{hand\_seals} \\
40--49      & 10  & \texttt{hand\_debuff} \\
50--59      & 10  & \texttt{selected} \\
\addlinespace
60--259     & 200 & \texttt{deck\_card\_ids} \\
260--459    & 200 & \texttt{deck\_enhancements} \\
460--659    & 200 & \texttt{deck\_editions} \\
660--859    & 200 & \texttt{deck\_seals} \\
860--1059   & 200 & \texttt{deck\_card\_locations} \\
1060--1259  & 200 & \texttt{deck\_debuff} \\
1260        & 1   & \texttt{deck\_total\_size} \\
1261        & 1   & \texttt{hand\_count} \\
\addlinespace
1262--1269  & 8   & \texttt{jokers} \\
1270--1277  & 8   & \texttt{joker\_editions} \\
1278--1285  & 8   & \texttt{joker\_sell\_values} \\
1286--1289  & 4   & \texttt{consumables} \\
1290--1293  & 4   & \texttt{consumable\_editions} \\
1294--1297  & 4   & \texttt{consumable\_sell\_values} \\
\addlinespace
1298        & 1   & \texttt{phase} \\
1299        & 1   & \texttt{ante} \\
1300        & 1   & \texttt{round\_number} \\
1301        & 1   & \texttt{blind\_type} \\
1302        & 1   & \texttt{round\_earned} \\
1303        & 1   & \texttt{boss\_blind\_id} \\
1304        & 1   & \texttt{small\_blind\_tag} \\
1305        & 1   & \texttt{big\_blind\_tag} \\
1306        & 1   & \texttt{double\_next\_tag} \\
1307        & 1   & \texttt{coupon\_tags} \\
1308        & 1   & \texttt{investment\_tags} \\
1309        & 1   & \texttt{payout\_investment} \\
1310        & 1   & \texttt{juggle\_tags} \\
\bottomrule
\end{tabular}
\end{minipage}
\hfill
\begin{minipage}[t]{0.48\textwidth}
\centering
\begin{tabular}{@{}lll@{}}
\toprule
\textbf{Range} & \textbf{Count} & \textbf{Field} \\
\midrule
1311--1320  & 10  & \texttt{pending\_tags} \\
1321--1330  & 10  & \texttt{queued\_tags} \\
\addlinespace
1331        & 1   & \texttt{chips\_scored} \\
1332        & 1   & \texttt{chips\_needed} \\
1333        & 1   & \texttt{hands\_left} \\
1334        & 1   & \texttt{discards\_left} \\
1335        & 1   & \texttt{money} \\
1336        & 1   & \texttt{pending\_round\_eval\_money} \\
1337--1348  & 12  & \texttt{hand\_levels} \\
\addlinespace
1349--1352  & 4   & \texttt{shop\_card\_item\_types} \\
1353--1356  & 4   & \texttt{shop\_card\_item\_ids} \\
1357--1360  & 4   & \texttt{shop\_card\_item\_editions} \\
1361--1364  & 4   & \texttt{shop\_card\_item\_enhancements} \\
1365--1368  & 4   & \texttt{shop\_card\_item\_seals} \\
1369--1370  & 2   & \texttt{shop\_pack\_items} \\
1371--1374  & 4   & \texttt{shop\_voucher\_items} \\
1375--1378  & 4   & \texttt{shop\_card\_prices} \\
1379--1380  & 2   & \texttt{shop\_pack\_prices} \\
1381--1384  & 4   & \texttt{shop\_voucher\_prices} \\
1385        & 1   & \texttt{reroll\_cost} \\
\addlinespace
1386--1390  & 5   & \texttt{pack\_item\_types} \\
1391--1395  & 5   & \texttt{pack\_item\_ids} \\
1396--1400  & 5   & \texttt{pack\_item\_editions} \\
1401--1405  & 5   & \texttt{pack\_item\_enhancements} \\
1406--1410  & 5   & \texttt{pack\_item\_seals} \\
\addlinespace
1411--1442  & 32  & \texttt{vouchers\_owned} \\
1443        & 1   & \texttt{idol\_card\_rank} \\
1444        & 1   & \texttt{idol\_card\_suit} \\
1445        & 1   & \texttt{ancient\_suit} \\
1446        & 1   & \texttt{mail\_rank} \\
1447        & 1   & \texttt{mail\_suit} \\
1448        & 1   & \texttt{castle\_suit} \\
\bottomrule
\end{tabular}
\end{minipage}
\end{table}

\end{document}